\documentclass[11pt]{article}
\usepackage[final]{acl}
\usepackage{times}
\usepackage{latexsym}
\usepackage[T1]{fontenc}
\usepackage[utf8]{inputenc}
\usepackage{microtype}
\usepackage{graphicx}
\usepackage{booktabs}
\usepackage{multirow}
\usepackage{textcomp}
\usepackage{amsthm}
\usepackage{amsmath}
\usepackage{amssymb}
\usepackage{enumitem}
\newtheorem{proposition}{Proposition}

\title{Belief Propagation in LLM World Models: Measuring Strategic Information Bias with Prediction Markets}

\author{
  \textbf{Mykola Khandoga}\textsuperscript{1},
  \textbf{Yevhen Kostiuk}\textsuperscript{1,2},
  \textbf{Anton Polishko}\textsuperscript{1},
  \textbf{Yurii Filipchuk}\textsuperscript{1}, \\
  \textbf{Kostiantyn Kozlov}\textsuperscript{1},
  \textbf{Dmytro Zamriy}\textsuperscript{1},
  \textbf{Artur Kiulian}\textsuperscript{1} \\
  \textsuperscript{1}Future Principle \quad
  \textsuperscript{2}Aarhus University
}

\begin{document}
\maketitle

\begin{abstract}
Every information ecosystem produces beliefs that shape strategic decisions. Both human analysts and AI systems inherit the blind spots of their information sources. We show that LLMs, combined with prediction markets, function as a calibrated instrument for measuring how far ecosystem-induced beliefs deviate from an external reference: LLMs extract the beliefs a text corpus implies, and prediction market price trajectories -- anchored at resolution by realised outcomes -- provide the calibration reference against which to quantify the deviation.

We isolate the bias contribution of specific text through ablation: varying information context while holding the model fixed, with a contaminated model that knows actual outcomes as control. Applied to 111 Ukraine-related prediction markets ({\raise.17ex\hbox{$\scriptstyle\sim$}}93,000 predictions, four models), we find that English news context systematically biases territorial predictions, wrong 64--72\% of the time ($p < 10^{-6}$). A contaminated model that knows actual outcomes shows the same error rate, indicating the bias originates primarily in the text. Supplementing with Ukrainian military-analytical sources reduces the bias for all clean models; absolute-error gains are partial and model-dependent.

We show that the distortion originates primarily in the sources, not the models. Consistent across four architectures, it will persist in any system that processes them and propagate into downstream decisions.
\end{abstract}

\section{Introduction}

The beliefs propagated by news coverage about ongoing events have major consequences for policy, public opinion, and resource allocation. Yet there exists no method to quantify how close they are to the public consensus. Existing approaches either detect framing properties of text without measuring their downstream cost \citep{ali2022survey, otmakhova2024framing}, or evaluate LLM forecasting accuracy without analyzing the information diet that drives it \citep{karger2025forecastbench, halawi2024approaching}.

Closing this gap requires solving two problems. First, we need a model that internalizes discourse framing -- not classifying frames from the outside, but absorbing them so its output reflects the belief the text induces. Second, we need a grounded scale against which to measure that belief -- an external reference anchored by realised outcomes, not another model's opinion.

LLMs solve the first problem. In-context learning operates as implicit Bayesian inference over latent concepts in the input \citep{xie2022explanation}, mechanistically equivalent to gradient descent on internal representations \citep{vonoswald2023transformers}. The model doesn't just read the text -- it updates its beliefs toward what the text implies. The output probability is the induced belief.

Prediction markets solve the second. A belief without an external reference is just an opinion. Prediction markets \citep{wolfers2004prediction,wolfers2006interpreting} provide continuous, financially incentivized probability estimates that are eventually anchored by realised outcomes. We use them in two distinct roles: the binary resolution gives us a low-power but genuine ground truth (§4.1), while the continuous price trajectory serves as the calibration reference for all model comparisons. The delta between the LLM-induced belief and market price, measured in percentage points (pp), is our calibrated measure of framing cost.

Our goal is to measure the bias that a specific text corpus induces -- how many pp closer to or further from the market estimate does this text push the model's prediction? An LLM prediction reflects both parametric knowledge from pretraining and in-context beliefs induced by the prompt. To isolate each source's contribution, we construct an ablation ladder:
\begin{itemize}[leftmargin=*, itemsep=1pt, topsep=2pt]
\item \textbf{A} provides market overview and current price data -- the minimal context from which the model reasons using only parametric knowledge.
\item \textbf{B} adds a price chart: structured numerical signal without narrative framing.
\item \textbf{C} adds English news articles: narrative without the rest of the context.
\item \textbf{D} combines all English-language sources -- chart, news, war map, trader comments -- representing the full English information ecosystem.
\item \textbf{D\textsubscript{UA}} supplements D with Ukrainian military sources: General Staff reports, frontline bloggers, defense media.
\end{itemize}
Each transition is a measurement in pp: A$\to$B measures the value of enriched price history signal, A$\to$C the cost of English news narrative alone, A$\to$D the cost of the full English ecosystem, D$\to$D\textsubscript{UA} the value of Ukrainian source diversification. A formal framework is given in Appendix~\ref{app:theory}; Appendix~\ref{app:decomposition} reports an exploratory decomposition of parametric and context-induced bias components.

Our contributions are:
\begin{enumerate}[leftmargin=*, itemsep=2pt, topsep=4pt]
    \item A method for measuring the bias a text corpus induces via belief propagation in LLMs, in calibrated probability units, validated against prediction markets;
    \item An ablation structure that isolates parametric from context-induced bias, revealing that the English information ecosystem systematically distorts LLM predictions on territorial questions -- a finding confirmed by a contaminated model control and by linguistic analysis of reasoning traces (offense-dominant verb framing, asymmetric counterfactual reasoning) -- and that supplementing with Ukrainian military-analytical sources partially counterbalances the bias;
    \item A dataset of {\raise.17ex\hbox{$\scriptstyle\sim$}}93,000 predictions with full reasoning traces.\footnote{\url{https://huggingface.co/datasets/OpenBabylon/unlp-ukraine-forecasting}}
\end{enumerate}

\section{Related Work}

\textbf{LLMs as forecasters.} ForecastBench \citep{karger2025forecastbench} shows LLMs now outperform non-expert crowds; agentic systems have reached superforecaster-level performance \citep{alur2025aia}. We depart from this line entirely: rather than benchmarking accuracy, we exploit LLMs' sensitivity to linguistic framing as a measurement instrument. Our models need not be good forecasters -- they need to faithfully reflect what the text implies.

\textbf{Computational framing analysis.} Media framing shapes public understanding of conflict \citep{entman2004projections}, and NLP work on factuality and bias of news media is surveyed in \citet{nakov2024survey}. Methods range from codebook annotation \citep{card2015media} to LLM-based classification of political bias in conflict coverage \citep{baly2020bias,alharbi2026evaluation}. Offense-dominant framing in Western coverage of Ukraine has been documented qualitatively \citep{ojala2024framing} and through topic modeling \citep{ptaszek2024war}. These approaches classify frames -- detecting what framing exists. Our method measures what that framing costs, in calibrated probability units.

\textbf{LLM bias.} Biases in LLMs are documented across demographic dimensions \citep{gallegos2024bias,feng2023pretraining}. These treat the model as the object of study. Our method measures how biased text propagates \emph{through} models into calibrated beliefs -- the model is the instrument, not the subject.

\section{Data and Methods}

\subsection{Dataset}

We collected 111 Ukraine-related prediction markets from Polymarket\footnote{\url{https://polymarket.com}} (January 2025 -- January 2026): 65 territorial (``Will side X control [city] by [date]?'') and 46 diplomatic (negotiations, sanctions, Zelenskyy suit). For each market, we constructed rolling prediction points at {\raise.17ex\hbox{$\scriptstyle\sim$}}8 cutoff dates with a median gap of {\raise.17ex\hbox{$\scriptstyle\sim$}}16 days, creating instances where we know the current price, the actual price at each horizon (6h, 12h, 1d, 2d, 3d, 5d and 7d), and all information available up to the cutoff.

\begin{figure}[t]
\centering
\includegraphics[width=\columnwidth]{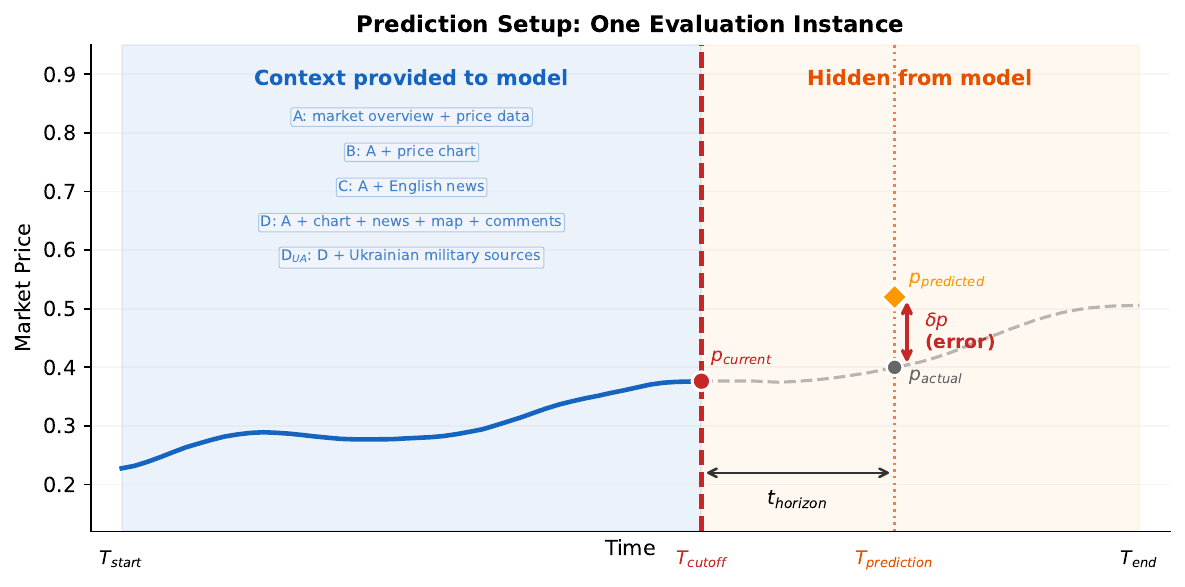}
\caption{One evaluation instance. The model receives context up to $T_{\mathrm{cutoff}}$ under varying information conditions (A--D\textsubscript{UA}); we compare its prediction at $T_{\mathrm{prediction}}$ against the actual market price.}
\label{fig:schematic}
\end{figure}

Each prediction is made under five information conditions: \textbf{A} (market overview + current price data only), \textbf{B} (A + price chart image), \textbf{C} (A + English news blocks), \textbf{D} (A + chart + English news + war map + Polymarket trader comments), and \textbf{D\textsubscript{UA}} (D supplemented with Ukrainian-language military sources: General Staff casualty reports, Telegram military bloggers, Militarnyi). Conditions are identical across models; the contaminated model (Gemini 3.1 Pro Preview) runs all conditions except D\textsubscript{UA}.

The English news corpus linked to our 111 benchmark markets comprises 16,457 articles from 2,217 domains (subset of a 122,290-article GDELT collection; Appendix~\ref{app:sources}): 89\% Western media, 3.4\% think tanks (ISW), 2.1\% Ukrainian English-language outlets, 2.8\% Russian sources, and zero Ukrainian-language analytical sources. The D\textsubscript{UA} corpus draws from DeepState frontline maps, General Staff daily loss reports, Ukrainian military bloggers, Militarnyi, Defense Express, and Ukrainian OSINT channels (Table~\ref{tab:ua_sources}).

\subsection{Models}

We evaluate three clean models -- Gemini 2.5 Flash, Gemini 2.5 Pro, and GPT-5-mini -- plus one contaminated control: Gemini 3.1 Pro Preview, whose training data extends past market resolution dates. The contaminated model's blind predictions show near-zero bias (+0.35\,pp vs +2.0\,pp clean average) and beat the no-change baseline by 10.4\%, confirming knowledge of actual outcomes.

Flash and Pro share training data (Gemini 2.5 family) but differ in reasoning depth, enabling controlled comparison of processing effects. All models output structured predictions with full reasoning traces.

\subsection{Statistical Framework}

The market is the unit of independence throughout ($N{=}65$ territorial, $N{=}46$ diplomatic). Adjacent cutoffs have overlapping prediction windows ({\raise.17ex\hbox{$\scriptstyle\sim$}}44\% overlap), so all tests use market-level aggregates with cluster-robust inference. We apply Bonferroni correction across 8 primary tests. With 65 clusters, we detect Cohen's $d \geq 0.35$ at 80\% power. Our accuracy baseline is \emph{no-change}: predict that the price at the horizon equals the current price. We use two distinct references: realised binary outcomes as the ground truth anchor in §4.1 (low-power but genuine), and the continuous market price trajectory as the calibration reference for all model comparisons in §4.2--4.4. Bias and MAE are therefore deviations from the market reference, not claims about reality. Full statistical details in Appendix~\ref{app:stats}.

\section{Results}

\subsection{The Benchmark Is Biased -- But Useful}

Polymarket overestimates Russian territorial capture by +3.5\,pp relative to actual outcomes ($t(381) = 5.58$, $p < 10^{-7}$, $d = 0.29$; this test is point-level since it characterises a property of the benchmark itself, distinct from the market-clustered tests used for model comparisons in §4.2--4.4). Despite this bias, the market correlates strongly with outcomes ($r = 0.83$, $R^2 = 0.69$), and binary resolution provides insufficient power (3/65 territorial markets resolved YES). Note that Polymarket resolution for territorial markets relies on ISW and DeepState frontline updates, themselves expert assessments rather than direct observation; residual disagreements between these sources and on-the-ground reality are absorbed into the +3.5\,pp figure. We use the continuous price trajectory as the calibration reference for all downstream comparisons.

\subsection{English Context Destroys Signal}

Blind models (condition A) show +2.0\,pp average pro-capture bias -- already present in training data, but \emph{less} than Polymarket's +3.5\,pp. Adding English news (condition D) pushes models to +3.4\,pp, nearly matching the market's overestimation. English news does not add information -- it replaces the model's more accurate prior with the discourse's less accurate framing. The shift is significant for Pro~2.5 (+2.4\,pp, $p = 0.0001$) and GPT-5-mini (+1.4\,pp, $p = 0.002$), both surviving Bonferroni correction. Flash shows a consistent but smaller effect (+0.5\,pp, $p = 0.066$). All bias measurements below are relative to the market price trajectory; \S4.1 establishes that this proxy is biased (+3.5\,pp) but strongly correlated with outcomes ($r = 0.83$), meaning our estimates are conservative -- the true distortion relative to reality is likely larger.

When context pushes predictions toward capture, those pushes are wrong 64--72\% of the time across all clean models (binomial test vs 50\%, $p < 10^{-6}$; Table~\ref{tab:push}). The bias is systematic and traceable: linguistic analysis of reasoning traces (Appendix~\ref{app:linguistic}) shows Russia receives 2.3--3.7$\times$ more offensive verbs, ``advances into'' territory 77 times while Ukraine never does, and models reason about ``if Russia succeeds'' 60 times but ``if Ukraine succeeds'' zero times. Bias$^2$ accounts for only 2--9\% of total error (Appendix~\ref{app:biasvar}); we study it because it is directional, not because it dominates accuracy.

\begin{figure}[t]
\centering
\includegraphics[width=\columnwidth]{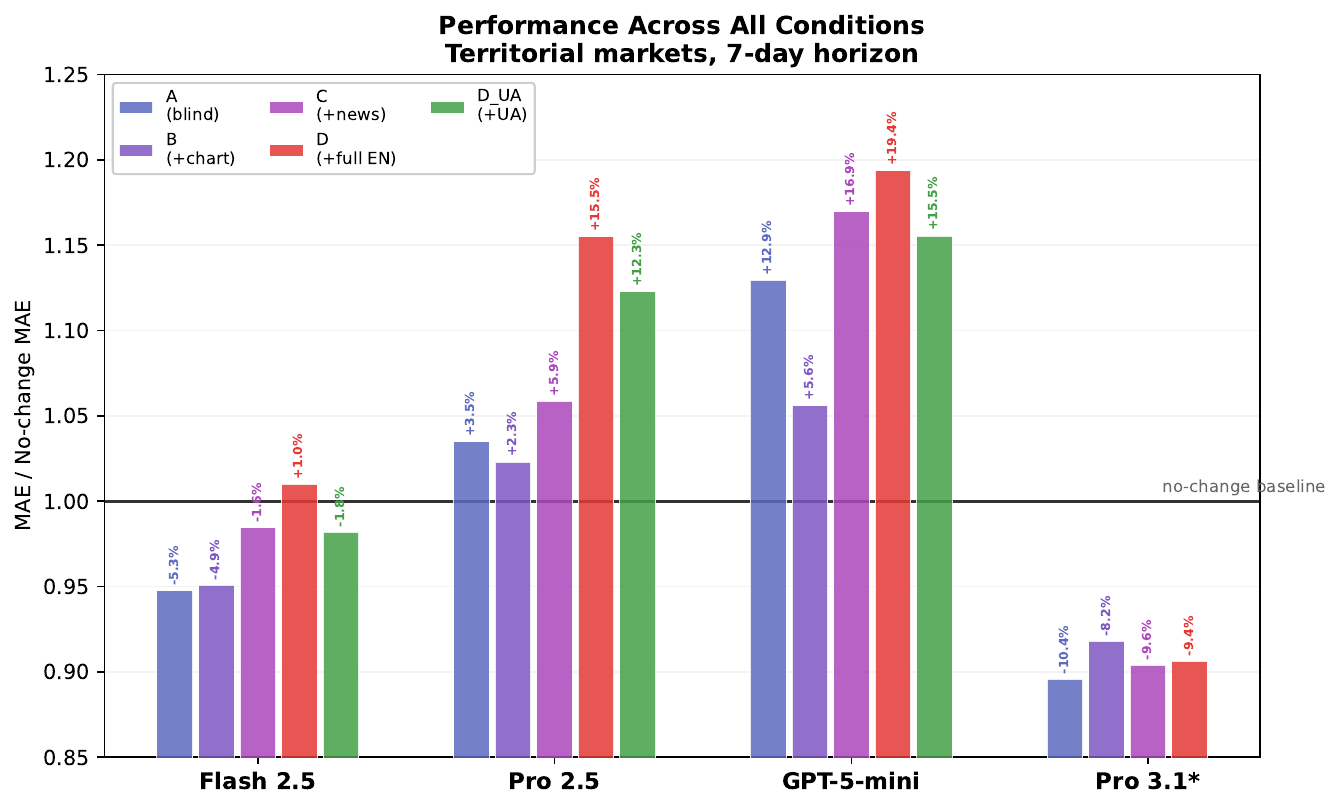}
\caption{MAE vs no-change baseline across all information conditions. Adding English context (A$\to$D) degrades predictions for all clean models; chart-only (B) outperforms full context (D). D\textsubscript{UA} partially recovers accuracy. The contaminated model (Pro~3.1*) beats baseline under all conditions; D\textsubscript{UA} was not run on it. Territorial markets, 7-day horizon.}
\label{fig:condition_ladder}
\end{figure}

The full condition ladder (Figure~\ref{fig:condition_ladder}; Appendix~\ref{app:conditions}) reveals that chart-only predictions (B) outperform full English context (D) for all clean models. Diplomatic markets serve as placebo: context does not damage predictions ($p = 0.03$ improvement for Flash), confirming a domain-specific mechanism (Appendix~\ref{app:diplomatic}). The bias is directionally asymmetric: downward predictions carry genuine signal, while upward predictions carry the discourse's offense-dominant distortion. Filtering out upward predictions converts all three models from losing to beating the no-change baseline (Appendix~\ref{app:selective}).

\subsection{Contaminated Model Ablation}

The contaminated model (Pro~3.1*) knows actual outcomes -- it beats no-change by 10.4\%. Yet when English context pushes it toward capture, it is wrong at the same rate as clean models (Table~\ref{tab:push}).

\begin{table}[t]
\centering
\small
\begin{tabular}{lcc}
\toprule
\textbf{Model} & \textbf{Knows?} & \textbf{Push$\uparrow$ acc.} \\
\midrule
Pro 3.1* (contam.) & Yes & 27.9\% \\
Flash 2.5 & No & 28.1\% \\
GPT-5-mini & No & 29.1\% \\
Pro 2.5 & No & 36.3\% \\
\bottomrule
\end{tabular}
\caption{When English context pushes predictions toward capture, accuracy is 28--36\% across four models with different knowledge levels ($p < 10^{-6}$ for all, binomial test vs 50\%). The error rate is a property of the corpus, not model ignorance.}
\label{tab:push}
\end{table}

The pro-capture push error rate is a property of the English news corpus, not of model ignorance (Appendix~\ref{app:contaminated}). An inverted-question robustness check confirms this is not an artifact of question framing (Appendix~\ref{app:antix}).

\subsection{Ukrainian Sources Reduce Bias; Accuracy Effects Are Model-Dependent}

\begin{figure}[t]
\centering
\includegraphics[width=\columnwidth]{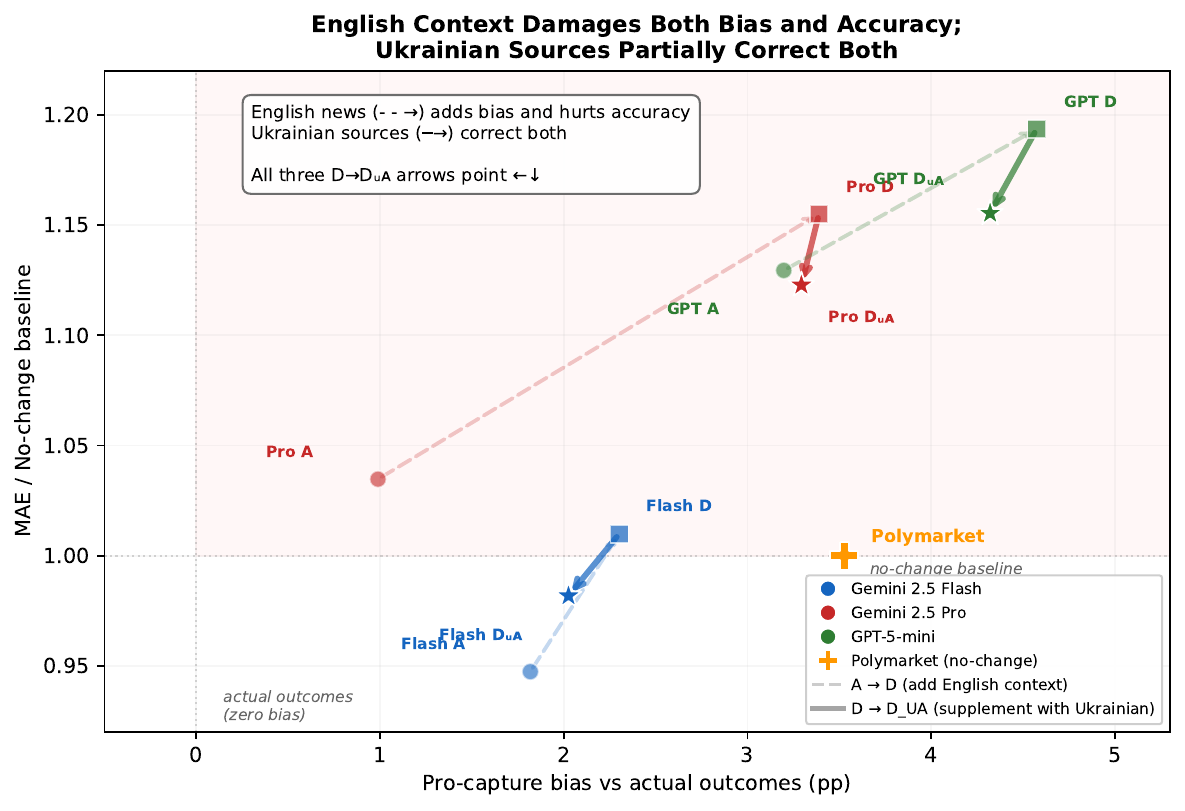}
\caption{Bias vs accuracy for each model under conditions A (blind), D (English context), and D\textsubscript{UA} (D supplemented with Ukrainian sources). Dashed arrows: A$\to$D (English context damages). Solid arrows: D$\to$D\textsubscript{UA} (Ukrainian sources correct). All solid arrows point left and down. Territorial markets, 7-day horizon.}
\label{fig:bias_accuracy}
\end{figure}

Supplementing English news with Ukrainian military sources (D $\to$ D\textsubscript{UA}) reduces pro-capture bias across all three clean models (Figure~\ref{fig:bias_accuracy}). We use ``correction'' here in the sense of reducing deviation from the market reference; D\textsubscript{UA} is source diversification, not fact-checking against a ground truth.

For Flash, D\textsubscript{UA} eliminates the context-induced directional shift: bias drops from +0.4\,pp ($p = 0.066$) to +0.1\,pp ($p = 0.378$), indistinguishable from blind prediction. For Pro and GPT, significant pro-capture bias persists ($p < 0.01$). Ukrainian sources provide a comparable absolute correction across models, but the outcome differs because models accumulate different amounts of context damage: Flash takes little damage from English text, so the correction is sufficient; Pro amplifies the offense-dominant signal far beyond what source supplementation can repair (Appendices~\ref{app:diplomatic},~\ref{app:horizons},~\ref{app:time_remaining}).

Accuracy effects are more mixed. D\textsubscript{UA} improves MAE on average across clean models, but not uniformly: the blind condition A remains competitive for some configurations, and D\textsubscript{UA} hurts diplomatic predictions (Appendix~\ref{app:diplomatic}). The robust finding is bias reduction; absolute error improvements are partial and model-dependent.

\section{Discussion and Conclusion}

Our results indicate a measurable cost of information ecosystem misalignment with the market reference: English-language context induces a systematic pro-capture bias through both biased framing and source exclusion, and Ukrainian military sources partially counterbalance it. The bias is robust ($p < 10^{-6}$ for push accuracy), domain-specific (territorial but not diplomatic), and invariant to model knowledge (contaminated model shows same push error rate).

The practical implications are immediate. Supplementing English retrieval with Ukrainian sources through multilingual RAG reduces directional bias across all clean models; absolute-error improvements follow on average but are partial and model-dependent. Conservative reasoning (Flash) benefits most, while deeper reasoning (Pro) amplifies the offense-dominant signal beyond what source supplementation can repair. This makes model selection a critical lever: choosing conservative over deep reasoning can matter more than improving the information itself. The parametric bias already present in condition~A reflects an English-centric information ecosystem and requires broadening source inclusion.

A case study illustrates the thesis in miniature: all clean models confidently predicted Zelenskyy would wear a suit to a papal funeral -- logically sound but culturally blind (Appendix~\ref{app:suit}). The method we introduce turns ecosystem misalignment from a qualitative concern into a grounded quantity.

\section*{Limitations}

With 65 territorial market clusters, we achieve 80\% power to detect Cohen's $d \geq 0.35$. Our MAE effects ($d = 0.25$--$0.31$) fall below this threshold, explaining non-significance after Bonferroni correction. The D\textsubscript{UA} vs D improvement does not reach significance at the market-cluster level ($p = 0.10$--$0.37$). The Flash/Pro processing divergence, while controlled (same training data), is a single comparison. Polymarket's Ukraine markets may not generalize to other conflicts. While the case study uses Ukrainian sources, the methodology generalises to any conflict where one information ecosystem is suspected of systematic framing distortion relative to another.

\section*{Ethical Considerations}

Our dataset spans 111 prediction markets about Ukraine -- from whether specific cities will be captured and communities displaced, to diplomatic negotiations, international sanctions, and aid decisions. The territorial markets reduce the fate of real communities to price movements on a trading platform. We find this commodification of human catastrophe deeply troubling.

We use this data not to legitimize prediction markets, but because their structure inadvertently exposes something important: a measurable gap between what the English-language information ecosystem implies about events in Ukraine and what is actually happening. This gap -- driven by both offense-dominant framing within included sources and exclusion of Ukrainian analytical ones -- has consequences beyond prediction accuracy. Distorted understanding shapes international policy, humanitarian response, and the political will to support Ukraine.

The induced bias is not merely dovish or negotiation-oriented. It systematically pushes toward a low-agency, offense-dominant view of Ukraine in which Ukrainian leverage is discounted and territorial loss is treated as more inevitable than reality later shows. The worldview induced by the Anglo-American source ecosystem qualitatively resembles later concessionist policy rhetoric that downplays Ukrainian leverage. We leave this striking alignment for future study.

The bias we document is not abstract. When English-language AI systems systematically overestimate Russian territorial success, they reinforce a narrative of inevitable Ukrainian loss that Ukrainian soldiers, analysts, and journalists work daily to counter -- through sources that English-language pipelines do not include.

This work used AI-based writing assistance tools for editing and formatting.

\section*{Acknowledgments}

We gratefully acknowledge Amazon Web Services and DigitalOcean for the cloud compute credits and infrastructure that supported this work. Model inference, training, and evaluation pipelines were run on AWS (EC2 GPU instances, SageMaker, S3); data collection and experiment tracking were hosted on DigitalOcean managed database and compute instances.

\bibliography{references}

\begin{thebibliography}{19}
\providecommand{\natexlab}[1]{#1}

\bibitem[{Al-Harbi et~al.(2026)}]{alharbi2026evaluation}
Mohammed Al-Harbi and 1 others. 2026.
\newblock An evaluation of {LLMs} for political bias in {Western} media:
  {Israel-Hamas} and {Ukraine-Russia} wars.
\newblock \emph{arXiv preprint arXiv:2601.06132}.

\bibitem[{Ali and Hassan(2022)}]{ali2022survey}
Mohammad Ali and Naeemul Hassan. 2022.
\newblock A survey of computational framing analysis approaches.
\newblock In \emph{Proceedings of the 2022 Conference on Empirical Methods in
  Natural Language Processing}, pages 9335--9348.

\bibitem[{Alur et~al.(2025)Alur, Stadie, Kang, Chen, McManus, Rickert, Lee,
  Federici et~al.}]{alur2025aia}
Rohan Alur, Bradly~C Stadie, Daniel Kang, Ryan Chen, Matt McManus, Michael
  Rickert, Tyler Lee, Michael Federici, and 1 others. 2025.
\newblock {AIA} forecaster: Technical report.
\newblock \emph{arXiv preprint arXiv:2511.07678}.

\bibitem[{Baly et~al.(2020)Baly, Da~San~Martino, Glass, and
  Nakov}]{baly2020bias}
Ramy Baly, Giovanni Da~San~Martino, James Glass, and Preslav Nakov. 2020.
\newblock We can detect your bias: Predicting the political ideology of news
  articles.
\newblock In \emph{Proceedings of the 2020 Conference on Empirical Methods in
  Natural Language Processing}, pages 4982--4991.

\bibitem[{Card et~al.(2015)Card, Boydstun, Gross, Resnik, and
  Smith}]{card2015media}
Dallas Card, Amber~E Boydstun, Justin~H Gross, Philip Resnik, and Noah~A Smith.
  2015.
\newblock The media frames corpus: Annotations of frames across issues.
\newblock In \emph{Proceedings of the 53rd Annual Meeting of the Association
  for Computational Linguistics}, pages 438--444.

\bibitem[{Entman(2004)}]{entman2004projections}
Robert~M Entman. 2004.
\newblock \emph{Projections of Power: Framing News, Public Opinion, and {U.S.}
  Foreign Policy}.
\newblock University of Chicago Press.

\bibitem[{Feng et~al.(2023)Feng, Park, Liu, and Tsvetkov}]{feng2023pretraining}
Shangbin Feng, Chan~Young Park, Yohan Liu, and Yulia Tsvetkov. 2023.
\newblock From pretraining data to language models to downstream tasks:
  Tracking the trails of political biases leading to unfair {NLP} models.
\newblock In \emph{Proceedings of the 61st Annual Meeting of the Association
  for Computational Linguistics}.

\bibitem[{Gallegos et~al.(2024)Gallegos, Rossi, Barber, Tanjim, Kim,
  Dernoncourt, Bui, Kim, Nushi, Yu et~al.}]{gallegos2024bias}
Isabel~O Gallegos, Ryan~A Rossi, Joe Barber, Md~Mehrab Tanjim, Sungchul Kim,
  Franck Dernoncourt, Tong Bui, Cheonbok Kim, Besmira Nushi, Duen~Horng Yu, and
  1 others. 2024.
\newblock Bias and fairness in large language models: A survey.
\newblock \emph{Computational Linguistics}, 50(3).

\bibitem[{Halawi et~al.(2024)Halawi, Shi, Borgeaud, Lerer, Abbeel, and
  Steinhardt}]{halawi2024approaching}
Danny Halawi, Fred Shi, Sebastian Borgeaud, Adam Lerer, Pieter Abbeel, and
  Jacob Steinhardt. 2024.
\newblock Approaching human-level forecasting with language models.
\newblock \emph{arXiv preprint arXiv:2402.18563}.

\bibitem[{Karger et~al.(2025)Karger, Bastani, Yueh-Han, Jacobs, Halawi, Zhang,
  and Tetlock}]{karger2025forecastbench}
Ezra Karger, Houtan Bastani, Chen Yueh-Han, Zachary Jacobs, Danny Halawi, Fred
  Zhang, and Philip~E Tetlock. 2025.
\newblock {ForecastBench}: A dynamic benchmark of {AI} forecasting
  capabilities.
\newblock In \emph{International Conference on Learning Representations}.

\bibitem[{Kuhn et~al.(2023)Kuhn, Gal, and Farquhar}]{kuhn2023semantic}
Lorenz Kuhn, Yarin Gal, and Sebastian Farquhar. 2023.
\newblock Semantic uncertainty: Linguistic invariances for uncertainty
  estimation in natural language generation.
\newblock In \emph{International Conference on Learning Representations}.

\bibitem[{Nakov et~al.(2024)Nakov, An, Kwak, Mansurov, and
  Mansurov}]{nakov2024survey}
Preslav Nakov, Jisun An, Haewoon Kwak, Muhammad~Arslan Mansurov, and Momin
  Mansurov. 2024.
\newblock A survey on predicting the factuality and the bias of news media.
\newblock In \emph{Findings of the Association for Computational Linguistics:
  ACL 2024}. Association for Computational Linguistics.

\bibitem[{Ojala et~al.(2024)Ojala, Pantti, and Kangas}]{ojala2024framing}
Markus Ojala, Mervi Pantti, and Jenni Kangas. 2024.
\newblock Framing the war in {Ukraine}: A comparative study of news coverage.
\newblock \emph{Journalism Studies}.

\bibitem[{Otmakhova et~al.(2024)Otmakhova, Khanehzar, and
  Frermann}]{otmakhova2024framing}
Yulia Otmakhova, Shima Khanehzar, and Lea Frermann. 2024.
\newblock Media framing: A typology and survey of computational approaches
  across disciplines.
\newblock In \emph{Proceedings of the 62nd Annual Meeting of the Association
  for Computational Linguistics (Volume 1: Long Papers)}.

\bibitem[{Ptaszek et~al.(2024)Ptaszek, Yuskiv, and Khomych}]{ptaszek2024war}
Grzegorz Ptaszek, Bogdan Yuskiv, and Serhii Khomych. 2024.
\newblock War on frames: Text mining of conflict in {Russian} and {Ukrainian}
  news agency coverage on {Telegram}.
\newblock \emph{Media, War \& Conflict}, 17(1):41--61.

\bibitem[{von Oswald et~al.(2023)von Oswald, Niklasson, Randazzo, Sacramento,
  Mordvintsev, Zhmoginov, and Vladymyrov}]{vonoswald2023transformers}
Johannes von Oswald, Eyvind Niklasson, Ettore Randazzo, Jo{\~a}o Sacramento,
  Alexander Mordvintsev, Andrey Zhmoginov, and Max Vladymyrov. 2023.
\newblock Transformers learn in-context by gradient descent.
\newblock In \emph{International Conference on Machine Learning}, pages
  35151--35174.

\bibitem[{Wolfers and Zitzewitz(2004)}]{wolfers2004prediction}
Justin Wolfers and Eric Zitzewitz. 2004.
\newblock Prediction markets.
\newblock \emph{Journal of Economic Perspectives}, 18(2):107--126.

\bibitem[{Wolfers and Zitzewitz(2006)}]{wolfers2006interpreting}
Justin Wolfers and Eric Zitzewitz. 2006.
\newblock Interpreting prediction market prices as probabilities.
\newblock NBER Working Paper 12200, National Bureau of Economic Research.

\bibitem[{Xie et~al.(2022)Xie, Raghunathan, Liang, and Ma}]{xie2022explanation}
Sang~Michael Xie, Aditi Raghunathan, Percy Liang, and Tengyu Ma. 2022.
\newblock An explanation of in-context learning as implicit {Bayesian}
  inference.
\newblock In \emph{International Conference on Learning Representations}.

\end{thebibliography}

\appendix

\section{Theoretical Framework}
\label{app:theory}

We formalize the mechanism studied in this paper. The goal is not a general theory of conflict forecasting, but to make precise what we mean by \emph{offense-dominant} beliefs under source exclusion.

\paragraph{Latent battlefield state.}
Let $i$ index a market-event pair. Let $y_i \in \{0,1\}$ denote the realized outcome, where $y_i=1$ corresponds to offensive success on the target event. Each event has an unobserved latent state $z_i \in \mathbb{R}$ capturing the true degree of offensive feasibility.

\paragraph{Information ecosystems as biased signals.}
We consider two information ecosystems: $E$ (English-language) and $U$ (Ukrainian military-analytical). Each provides a noisy signal about the same latent state:
\[
x_i^E = z_i + \beta_E + \varepsilon_i^E,
\qquad
x_i^U = z_i + \beta_U + \varepsilon_i^U,
\]
where $\varepsilon_i^E, \varepsilon_i^U$ are zero-mean noise and $\beta_E, \beta_U \in \mathbb{R}$ are systematic shifts. An ecosystem is \emph{more offense-dominant} when it shifts beliefs further toward offensive success. The English ecosystem is more offense-dominant than the Ukrainian one whenever $\beta_E > \beta_U$.

\paragraph{LLMs as belief elicitation instruments.}
For model $m$, let $\mu_m$ denote the model's effective baseline belief under condition A (minimal context: market overview and current price data), absorbing both pretraining priors and the model's processing of the price signal. Given a source set $S$, the model produces
\[
\hat p_i^{(m)}(S) = \sigma\!\bigl(g_i^{(m)}(S)\bigr),
\]
where $\sigma(\cdot)$ is the logistic sigmoid and $g_i^{(m)}(S)$ is a latent score formed by combining the model prior with available signals:
\[
g_i^{(m)}(S)
=
\frac{
\lambda_{m,0}\mu_m + \sum_{e \in S}\lambda_{m,e}\, x_i^e
}{
\lambda_{m,0} + \sum_{e \in S}\lambda_{m,e}
}.
\]
Here $\lambda_{m,0} \geq 0$ weights the prior and $\lambda_{m,e} \geq 0$ weights ecosystem $e$. We treat these as effective parameters that absorb presentation-order effects and model-specific context processing. This maps onto our three experimental conditions:
\begin{align*}
\hat p_i^{(m)}(A) &= \sigma(\mu_m), \\
\hat p_i^{(m)}(D) &= \hat p_i^{(m)}(\{E\}), \\
\hat p_i^{(m)}(D_{UA}) &= \hat p_i^{(m)}(\{E,U\}).
\end{align*}

\begin{proposition}[Source exclusion shifts latent scores toward offense]
\label{prop:source_exclusion}
Assume ecosystem signals satisfy $x_i^e = z_i + \beta_e + \varepsilon_i^e$ with $\mathbb{E}[\varepsilon_i^e]=0$, and that the latent score $g_i^{(m)}(S)$ is a positively weighted average of the model prior and available signals. If $\beta_E > \beta_U$ and $\lambda_{m,U}>0$, then for every event $i$:
\[
\mathbb{E}\!\left[g_i^{(m)}(D)\right]
>
\mathbb{E}\!\left[g_i^{(m)}(D_{UA})\right].
\]
Excluding Ukrainian sources systematically increases the expected latent score toward offensive success.
\end{proposition}

\paragraph{Proof.}
Substituting $\mathbb{E}[x_i^E]=z_i+\beta_E$ and $\mathbb{E}[x_i^U]=z_i+\beta_U$:
\[
\mathbb{E}\!\left[g_i^{(m)}(D)\right]
=
\frac{
\lambda_{m,0}\mu_m + \lambda_{m,E}(z_i+\beta_E)
}{
\lambda_{m,0}+\lambda_{m,E}
},
\]
\begin{align*}
&\mathbb{E}\!\left[g_i^{(m)}(D_{UA})\right] \\
&= \frac{\lambda_{m,0}\mu_m + \lambda_{m,E}(z_i+\beta_E) + \lambda_{m,U}(z_i+\beta_U)}{\lambda_{m,0}+\lambda_{m,E}+\lambda_{m,U}}.
\end{align*}
Since $\lambda_{m,U}>0$ and $\beta_U < \beta_E$, the additional term pulls the weighted average below the English-only value. \hfill $\square$

\paragraph{From scores to probabilities.}
Since $\sigma(\cdot)$ is monotone increasing, the score ordering implies $\hat p_i^{(m)}(D) > \hat p_i^{(m)}(D_{UA})$ pointwise for each realization of the noise. The ordering $\mathbb{E}[\hat p_i^{(m)}(D)] > \mathbb{E}[\hat p_i^{(m)}(D_{UA})]$ then follows by taking expectations. However, the \emph{magnitude} of the probability gap depends on the curvature of $\sigma$ and the noise distribution, so the proposition is stated at the level of latent scores where the result is exact.

\paragraph{Operational quantities.}
The framework yields three empirically measurable quantities. The \emph{harm of English context}: $H_m = \mathbb{E}_i[\hat p_i^{(m)}(D)-\hat p_i^{(m)}(A)]$. The \emph{source exclusion cost}: $X_m = \mathbb{E}_i[\hat p_i^{(m)}(D)-\hat p_i^{(m)}(D_{UA})]$. The \emph{directional bias}: $B_m(S) = \mathbb{E}_i[\hat p_i^{(m)}(S)-y_i]$. On territorial markets, positive $B_m(S)$ indicates systematic overprediction of offensive success. Our main empirical finding is $B_m(D) > B_m(D_{UA})$, with $H_m > 0$ and $X_m > 0$.

\paragraph{Limitations of the formalization.}
This framework is deliberately modest. It does not claim that LLM outputs recover latent beliefs perfectly, nor that ecosystems differ only along a single dimension. The weighted-average assumption is an idealization: actual LLMs process context sequentially, and effective weights may depend on presentation order, prompt structure, and reasoning depth. The formalization makes explicit the mechanism tested -- if one ecosystem is more offense-dominant, excluding the other should produce more offense-dominant predictions -- without claiming more than the experimental design supports.

\section{Statistical Methodology}
\label{app:stats}

\paragraph{Clustering.} Adjacent cutoffs within a market have overlapping 7-day prediction windows ({\raise.17ex\hbox{$\scriptstyle\sim$}}44\% overlap at the median 16-day gap). Intra-cluster correlation ranges from 0.008 (Flash) to 0.141 (GPT-5-mini). All tests use market-level aggregates.

\paragraph{Multiple testing.} We apply Bonferroni correction across 8 primary tests. Market bias (territorial and diplomatic), directional bias shift (3 models), and MAE damage (3 models). After correction: market bias ($p < 10^{-4}$), Pro~2.5 bias shift ($p = 6.8 \times 10^{-4}$), and GPT bias shift ($p = 0.018$) survive. MAE tests and Flash bias shift do not survive. Push accuracy tests ($p < 10^{-6}$) are excluded from this family.

\paragraph{Power.} With $N{=}65$ clusters, 80\% power at $\alpha = 0.05$ requires $d \geq 0.35$. Observed: Pro~2.5 $d = 0.50$ (powered), GPT $d = 0.36$ (marginal), Flash $d = 0.19$ (underpowered). MAE effects $d = 0.25$--$0.31$ (underpowered). Flash-scale detection requires {\raise.17ex\hbox{$\scriptstyle\sim$}}215 clusters.

\paragraph{Permutation tests.} Sign-randomization tests (10,000 iterations) confirm all parametric results with concordant $p$-values.

\section{Question Inversion (Anti-X)}
\label{app:antix}

We re-run territorial predictions with inverted questions: ``Will Russia \emph{fail to capture} X?'' with inverted prices ($1{-}p$). This is analogous to semantic entropy \citep{kuhn2023semantic} but diagnoses \emph{context quality} rather than model confidence.

When contradictions occur under condition D, they are directionally asymmetric. For Pro~3.1* (contaminated), the asymmetry is 47:1 -- original-down/anti-up vastly dominates. English military reporting is stronger evidence \emph{against failure} than \emph{for capture}.

\section{Selective Trust}
\label{app:selective}

A zero-parameter rule -- trust the model only when it predicts downward movement -- converts losing strategies to winning ones. Flash selective: $-$3.4\% vs no-change ($p < 10^{-21}$). Pro~2.5 selective: $-$3.3\%. This reveals that downward predictions carry genuine signal while upward (pro-capture) predictions carry systematic noise. D\textsubscript{UA} + selective is the best combination for GPT-5-mini ($-$0.8\%), finally converting it to a winning strategy.

\section{Per-Horizon Analysis}
\label{app:horizons}

\begin{figure}[t]
\centering
\includegraphics[width=\columnwidth]{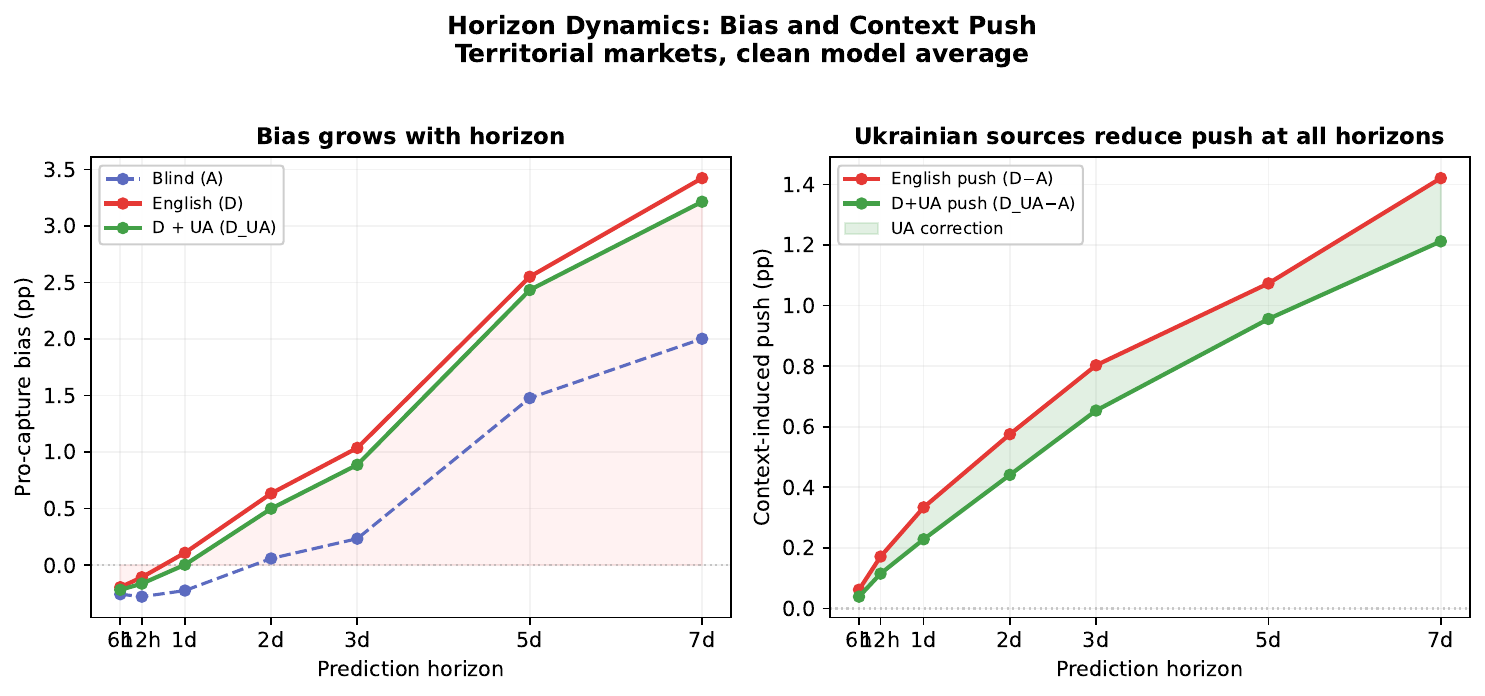}
\caption{Left: pro-capture bias grows with prediction horizon for all conditions. Right: context-induced push (D$-$A) and Ukrainian-corrected push (D\textsubscript{UA}$-$A) across horizons. D\textsubscript{UA} reduces push at every horizon. Territorial markets, clean model average.}
\label{fig:horizon_dynamics}
\end{figure}

Pro-capture bias grows with prediction horizon (Table~\ref{tab:horizons}, Figure~\ref{fig:horizon_dynamics}). D\textsubscript{UA} consistently produces less push than D at every horizon.

\begin{table}[t]
\centering
\small
\begin{tabular}{lccccc}
\toprule
\textbf{Hz} & \textbf{A} & \textbf{D} & \textbf{D\textsubscript{UA}} & \textbf{D push} & \textbf{D\textsubscript{UA} push} \\
\midrule
6h & $-$0.3 & $-$0.2 & $-$0.2 & +0.1 & +0.0 \\
1d & $-$0.2 & +0.1 & +0.0 & +0.3 & +0.2 \\
3d & +0.2 & +1.0 & +0.9 & +0.8 & +0.7 \\
7d & +2.0 & +3.4 & +3.2 & +1.4 & +1.2 \\
\bottomrule
\end{tabular}
\caption{Bias by horizon (pp, clean model average, territorial). Bias grows with horizon; D\textsubscript{UA} reduces push at all horizons.}
\label{tab:horizons}
\end{table}

\section{Source Ecosystem and Utility}
\label{app:sources}

\subsection{English Information Diet}

The full GDELT collection spans all topics and comprises 122,290 articles from 6,232 domains. Of these, 16,457 articles from 2,217 domains are linked to our 111 Ukraine benchmark markets (\S3). The composition statistics below describe the benchmark subset. The corpus is US-centric (63\% of articles) and Western-oriented (Table~\ref{tab:en_sources}).

\begin{table}[h]
\centering
\small
\begin{tabular}{llr}
\toprule
\textbf{Category} & \textbf{Examples} & \textbf{\%} \\
\midrule
Wire / mainstream & AP, Reuters, CNN, Guardian & 36 \\
Aggregators & Yahoo, AOL, bignewsnetwork & 12 \\
International & jpost, economictimes (India) & 12 \\
Conservative US & Fox, Breitbart, Epoch Times & 5 \\
Finance & marketscreener, fxstreet, Forbes & 5 \\
UK press & Daily Mail, Independent & 5 \\
Specialist / OSINT & ISW, globalsecurity.org & 1 \\
Other / regional & various & 24 \\
\bottomrule
\end{tabular}
\caption{English source composition by category. US sources account for 63\% of articles. Ukrainian-language analytical sources: zero.}
\label{tab:en_sources}
\end{table}

The top domains by volume are Yahoo (5,864), marketscreener (1,142), freerepublic (781), Daily Mail (736), globalsecurity.org (659), ZeroHedge (595), ABC News (560), ISW (534), Independent (464), and Newsweek (462). The corpus reflects what GDELT surfaces as the English-language information diet about Ukraine -- mainstream wire content, finance aggregators, and partisan outlets. \textbf{Zero Ukrainian-language analytical sources appear.}

\subsection{Ukrainian Information Diet}

The D\textsubscript{UA} condition supplements condition D with three source types (Table~\ref{tab:ua_sources}).

\begin{table}[h]
\centering
\small
\begin{tabular}{lrl}
\toprule
\textbf{Source} & \textbf{Count} & \textbf{Type} \\
\midrule
Telegram bloggers & 47,009 posts & 16 channels \\
Militarnyi.com & 2,352 articles & Defense news \\
General Staff losses & 1,468 records & Daily casualty data \\
\bottomrule
\end{tabular}
\caption{Ukrainian sources added in D\textsubscript{UA}.}
\label{tab:ua_sources}
\end{table}

The 16 Telegram channels represent the core of the Ukrainian military information ecosystem: 5 active/reserve military (including brigade commanders Magyar, Zhorin, Shtefan, Fedorenko, Krotevych), 4 journalists (Tsaplienko, Butusov, Kazanskyi), 4 analysts (Berezovets, Kovalenko, Mashovets, Zhdanov), 1 official channel (General Staff ZSU), 1 tech specialist (Beskrestnov), and 1 commentator. All channels represent Ukrainian perspectives -- no Russian, neutral, or Western-OSINT channels are included. This is by design: D\textsubscript{UA} tests whether adding the Ukrainian military-analytical ecosystem improves prediction accuracy, not whether balanced sourcing does.

\subsection{Source Utility Analysis}

We analyze which sources models actually cite in their reasoning traces (condition D, 7d horizon, clean models, $N = 2{,}130$ predictions) and whether citing a source correlates with better or worse predictions. The patterns below are correlational: ISW-cited predictions performing worse does not establish that citing ISW \emph{causes} worse predictions; both may reflect harder markets or more contested events. Relative performance compares the model's error when citing a source against the no-change baseline for those same markets.

\begin{table}[h]
\centering
\small
\setlength{\tabcolsep}{3pt}
\begin{tabular}{lrrrr}
\toprule
\textbf{Source} & \textbf{Freq} & \textbf{Rel.} & \textbf{Dir\%} & \textbf{Dir\%} \\
 & & \textbf{Perf} & \textbf{(cited)} & \textbf{(not)} \\
\midrule
Russian officials & 43.0\% & $-$1.4\% & \textbf{61.9} & 51.9 \\
Ukrainian officials & 36.9\% & $-$5.4\% & 60.0 & 54.0 \\
Price action & 76.9\% & $-$8.6\% & 56.7 & 54.5 \\
ISW & 51.5\% & $-$11.3\% & 50.9 & \textbf{61.8} \\
Russian milbloggers & 6.7\% & $-$11.9\% & 54.8 & 56.3 \\
Polymarket traders & 17.7\% & $-$6.5\% & 55.6 & 56.3 \\
DeepState & 3.8\% & $-$1.0\% & 56.1 & 56.2 \\
UA General Staff & 5.4\% & $-$15.1\% & 41.5 & 57.1 \\
OSINT & 1.9\% & $-$14.6\% & 41.5 & 56.5 \\
\bottomrule
\end{tabular}
\caption{Source utility: frequency of citation in reasoning traces and correlation with prediction quality. Relative performance = (NC\_error $-$ model\_error) / NC\_error; negative = worse than no-change. ISW -- the most-cited analytical source (51.5\%) -- correlates with \emph{worse} directional accuracy (50.9\%) than predictions not citing it (61.8\%).}
\label{tab:source_utility}
\end{table}

\paragraph{The ISW paradox.} ISW is the dominant analytical source (51.5\% citation rate). Yet predictions citing ISW show 50.9\% directional accuracy -- a coin flip -- compared to 61.8\% when ISW is not cited. This is consistent with ISW's offense-dominant analytical framing: detailed, authoritative coverage of offensive operations that models internalize as evidence for territorial change. The authority of the source amplifies the framing effect.

\paragraph{Russian officials as accidental signal.} Russian officials are cited at 43.0\% with the best directional accuracy (61.9\%) among high-frequency sources. This is counterintuitive until one considers the epistemic framing from Appendix~\ref{app:linguistic}: models treat Russian claims with skepticism (``Russia \emph{claims}...''), and this discounting accidentally produces better-calibrated predictions than absorbing ISW's authoritative framing uncritically.

\paragraph{Contaminated model source shift.} The contaminated model (Pro~3.1*), which knows outcomes, shifts its citation patterns: ISW drops from 51.5\% to 35.4\%, Ukrainian officials from 36.9\% to 23.1\%, while price action rises from 76.9\% to 81.0\%. A model with outcome knowledge relies less on narrative sources and more on the price signal -- further evidence that narrative sources add framing, not information.

\section{Bias-Variance Decomposition}
\label{app:biasvar}

Bias$^2$ accounts for 2--9\% of model MSE; variance dominates at 91--98\%. Polymarket: 8\% bias$^2$. English context increases both components; Ukrainian sources reduce both. Variance reduction is the larger contributor to MAE improvement for bold models.

\section{Three-Layer Bias Decomposition}
\label{app:decomposition}

\begin{figure}[t]
\centering
\includegraphics[width=\columnwidth]{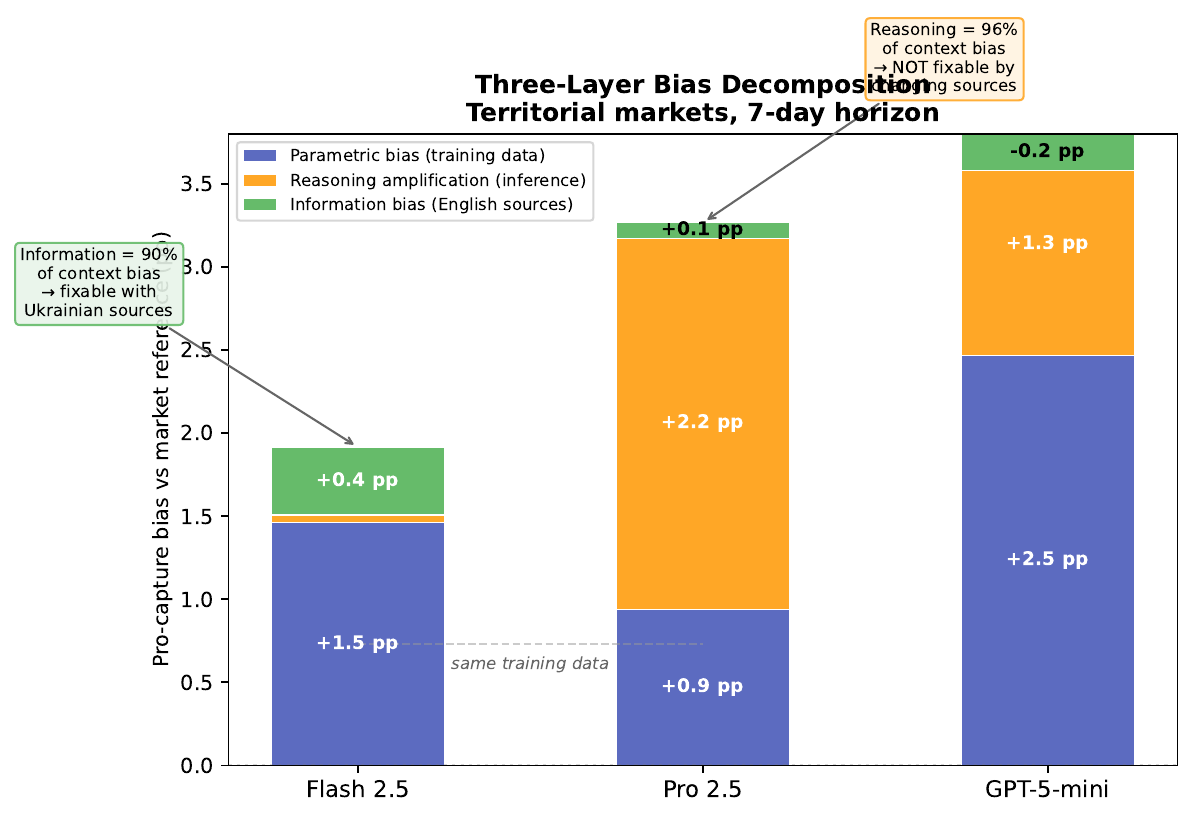}
\caption{Bias decomposition. Flash and Pro share training data but differ 5$\times$ in context-induced bias. Ukrainian sources provide a similar absolute correction for all models, but it is swamped by Pro's context damage.}
\label{fig:decomposition}
\end{figure}

We present this decomposition as exploratory and correlational. With 65 territorial markets, per-model component estimates are noisy and we do not draw causal source-level conclusions from them. Our experimental conditions separate three components of pro-capture bias (Figure~\ref{fig:decomposition}), measured on territorial markets at 7-day horizon.

\paragraph{Parametric bias (training data).} Measured by condition A -- no context, just the model's priors. Flash: +1.8\,pp. Pro: +1.0\,pp. GPT: +3.2\,pp.

\paragraph{Context-induced bias (English news).} Measured by D$-$A: the additional pro-capture shift from English context. This is where models diverge dramatically. Flash: +0.5\,pp. Pro: +2.4\,pp. GPT: +1.4\,pp. Pro accumulates 5$\times$ more context damage than Flash despite sharing training data -- deeper reasoning amplifies the offense-dominant signal in English text.

\paragraph{Ukrainian source correction.} Supplementing with Ukrainian sources provides a roughly constant absolute correction across models: Flash $-$0.3\,pp, Pro $-$0.1\,pp, GPT $-$0.3\,pp (Table~\ref{tab:decomposition}). What differs is not the correction but the denominator -- how much context damage each model accumulates. For Flash, 0.3\,pp corrects most of the 0.5\,pp context damage (57\%). For Pro, a similar correction is negligible against 2.4\,pp of damage (4\%).

\begin{table}[t]
\centering
\small
\begin{tabular}{lccc}
\toprule
\textbf{Model} & \textbf{Param.} & \textbf{Context} & \textbf{UA corr.} \\
\midrule
Flash 2.5 & +1.8\,pp & +0.5\,pp & $-$0.3\,pp \\
Pro 2.5 & +1.0\,pp & +2.4\,pp & $-$0.1\,pp \\
GPT-5-mini & +3.2\,pp & +1.4\,pp & $-$0.3\,pp \\
\bottomrule
\end{tabular}
\caption{Three-layer bias decomposition. Parametric bias (A) is in the weights. Context-induced bias (D$-$A) comes from English news. Ukrainian correction (D$-$D\textsubscript{UA}) is the bias reduction from supplementing with Ukrainian sources -- roughly constant across models, but against vastly different context damage.}
\label{tab:decomposition}
\end{table}

\begin{table}[t]
\centering
\small
\begin{tabular}{lcc}
\toprule
\textbf{Model} & \textbf{Bias recov.} & \textbf{Accuracy recov.} \\
\midrule
Flash 2.5 & 57\% & 45\% \\
Pro 2.5 & 4\% & 27\% \\
GPT-5-mini & 18\% & 60\% \\
\bottomrule
\end{tabular}
\caption{D\textsubscript{UA} recovery of context-induced damage. All models improve on both dimensions.}
\label{tab:recovery}
\end{table}

This is a model selection result: Ukrainian sources help all models roughly equally in absolute terms, but their impact depends on how aggressively the model amplifies English context. Conservative reasoning (Flash) keeps context damage small, making the correction sufficient. Deep reasoning (Pro) amplifies English bias beyond what source supplementation can repair.

\section{Full Condition Comparison}
\label{app:conditions}

\begin{table}[t]
\centering
\small
\begin{tabular}{lccccc}
\toprule
& \textbf{A} & \textbf{B} & \textbf{C} & \textbf{D} & \textbf{D\textsubscript{UA}} \\
\midrule
Flash & $-$5.3 & $-$4.9 & $-$1.6 & +1.0 & $-$1.8 \\
Pro & +3.5 & +2.3 & +5.9 & +15.5 & +12.3 \\
GPT & +12.9 & +5.6 & +16.9 & +19.4 & +15.5 \\
3.1* & $-$10.4 & $-$8.2 & $-$9.6 & $-$9.4 & -- \\
\bottomrule
\end{tabular}
\caption{MAE vs no-change (\%), territorial 7d. Chart-only (B) outperforms full context (D) for all clean models.}
\label{tab:conditions}
\end{table}

\begin{figure}[t]
\centering
\includegraphics[width=\columnwidth]{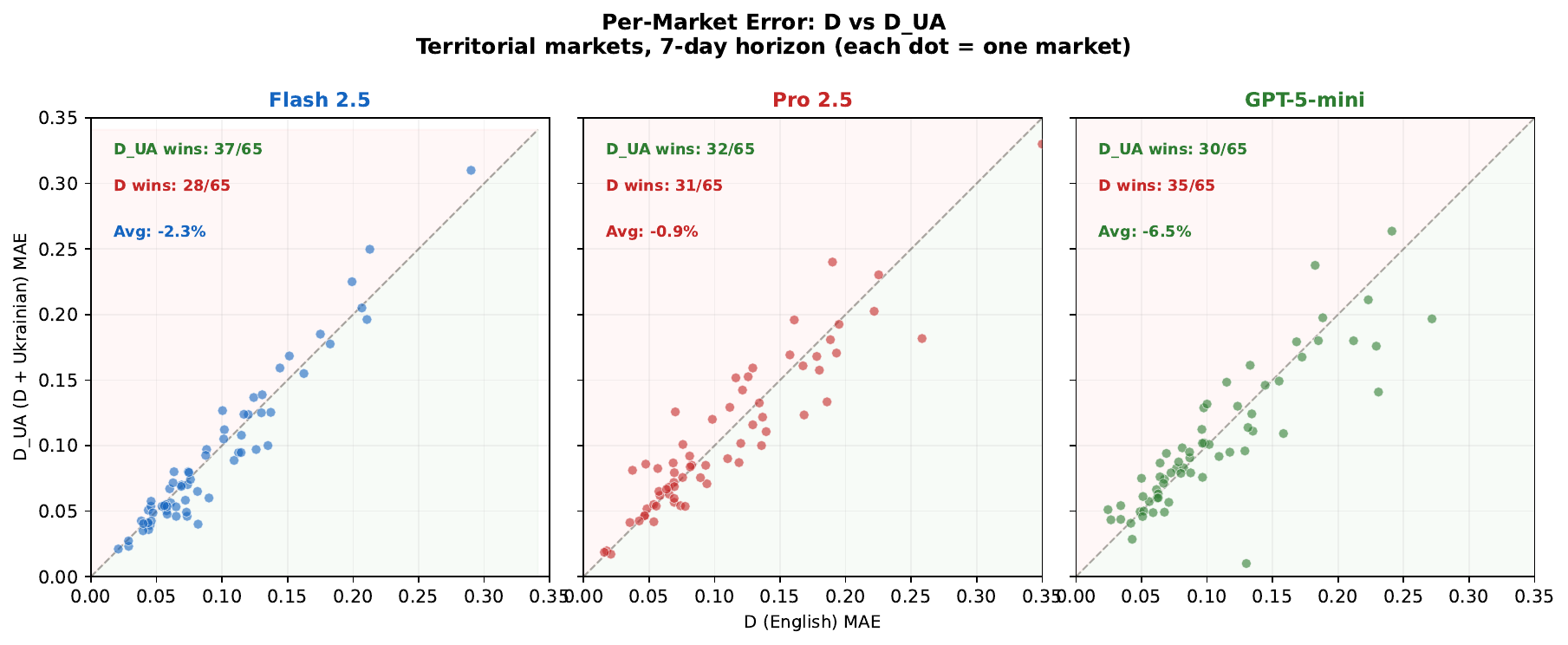}
\caption{Per-market MAE under D (English) vs D\textsubscript{UA} (supplemented with Ukrainian sources). Each dot is one territorial market. Points below the diagonal indicate D\textsubscript{UA} outperforms D. D\textsubscript{UA} wins the majority of markets for Flash (37/65) and Pro (32/65). Territorial markets, 7-day horizon.}
\label{fig:per_market}
\end{figure}

\section{Contaminated Model Details}
\label{app:contaminated}

Pro~3.1* shows near-zero blind bias (+0.35\,pp vs +2.0\,pp clean average) and beats no-change by 10.4\% blind, 9.4\% with context -- the only model to beat baseline under any condition. Despite this knowledge, its pro-capture push accuracy (27.9\%) matches clean models, demonstrating that the directional signal in English military reporting overrides even direct outcome knowledge.

\section{Diplomatic Markets (Placebo)}
\label{app:diplomatic}

\begin{figure}[t]
\centering
\includegraphics[width=\columnwidth]{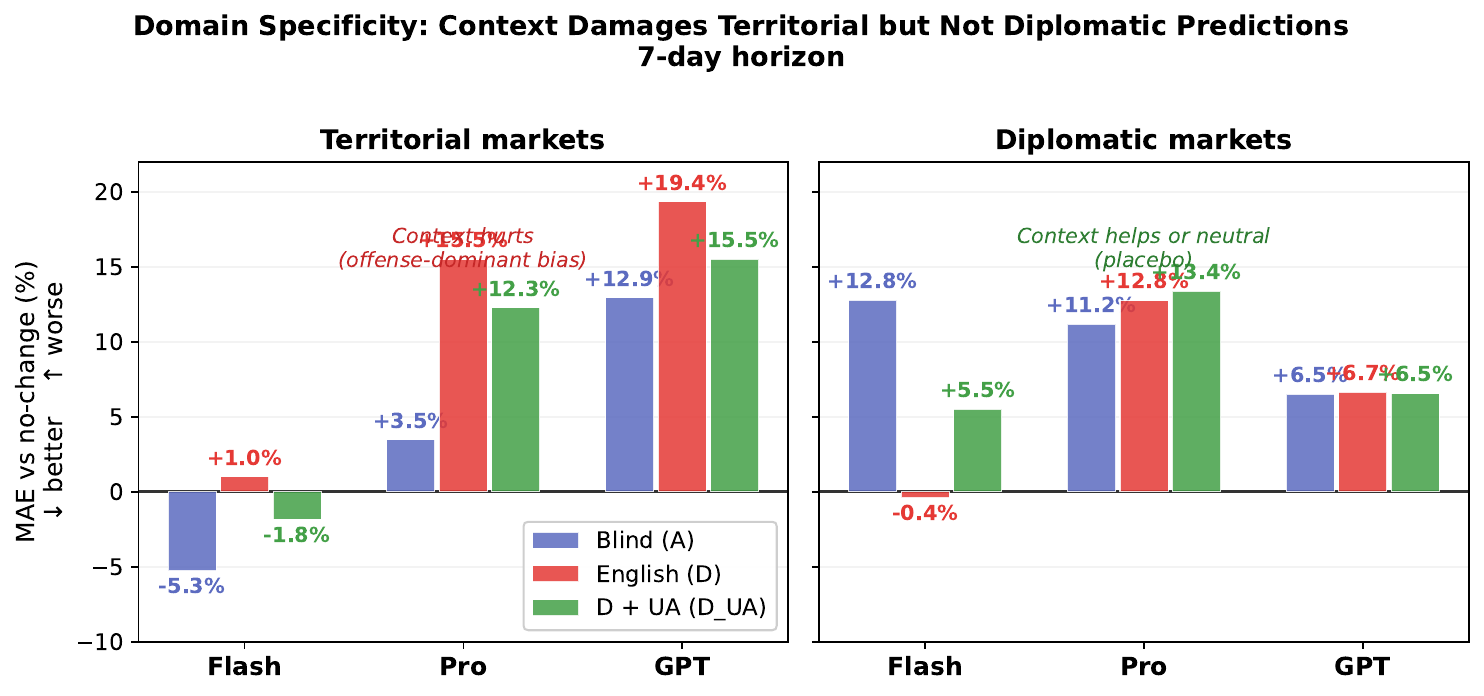}
\caption{Domain specificity: English context damages territorial predictions (left) but not diplomatic ones (right). D\textsubscript{UA} corrects territorial bias but hurts diplomatic predictions, confirming the intervention is domain-specific. 7-day horizon.}
\label{fig:territorial_diplomatic}
\end{figure}

D\textsubscript{UA} hurts diplomatic predictions for Flash (+5.9\% vs D) and Pro (+5.8\%), while GPT is unchanged ($-$0.1\%). Ukrainian military sources contain no diplomatic signal; English coverage of negotiations and sanctions is more informative. This confirms domain specificity (Figure~\ref{fig:territorial_diplomatic}).

\section{Linguistic Analysis of Offense-Dominant Framing}
\label{app:linguistic}

The quantitative results in \S4 stand independently of the analysis below. We provide linguistic analysis of reasoning traces as qualitative illustration of the mechanism behind the statistical findings.

We analyze 444 reasoning traces (111 markets $\times$ 4 models, condition D, 7d horizon, first cutoff per market; {\raise.17ex\hbox{$\scriptstyle\sim$}}159,000 words) using regex-based proximity matching with manual validation. The analysis reveals eight dimensions of offense-dominant framing, all pointing in the same direction across all models.

\subsection{Agency: Who Acts}

Russia is systematically framed as the agent. Across all models, Russia appears as grammatical subject 60--66\% of the time (ratio 1.4--2.0$\times$ over Ukraine), is the first actor mentioned in 76--86\% of texts, and receives 1.56--1.76$\times$ more total mentions. Ukraine's primary role is reactive.

\subsection{Verb Semantics: What Each Side Does}

Within 80 characters of each actor mention, 59--64\% of verbs near Russia are offensive (advance, capture, assault, deploy) while Ukraine's verbs split between defensive (hold, resist, repel) and diplomatic. Russia receives 2.3--3.7$\times$ more offensive verbs than Ukraine across all four models.

\subsection{Success/Failure Framing}

The most extreme asymmetry. Russia's success-to-failure language ratio ranges from 6.7:1 to 14.1:1. Ukraine's ratio is 1.4--2.5:1. Russia is described in almost exclusively positive-outcome terms (advance, progress, gains, momentum) even when the model predicts those outcomes will not materialize. The contaminated model (Pro~3.1*) shows the highest ratio (14.1:1) despite knowing most territorial outcomes resolve against capture -- framing and prediction are decoupled.

\subsection{Territorial Lexicon}

\begin{table}[t]
\centering
\small
\begin{tabular}{lccc}
\toprule
\textbf{Verb} & \textbf{Russia} & \textbf{Ukraine} & \textbf{Ratio} \\
\midrule
capture & 143 & 16 & 8.9$\times$ \\
advance into & 77 & \textbf{0} & $\infty$ \\
seize & 14 & 1 & 14$\times$ \\
encircle & 15 & 2 & 7.5$\times$ \\
hold & 2 & 3 & 0.7$\times$ \\
\bottomrule
\end{tabular}
\caption{Territorial verbs attributed to each actor (sum across 4 models, 444 traces). ``Advance into'' appears 77 times near Russia and \textbf{zero} times near Ukraine.}
\label{tab:territorial_verbs}
\end{table}

\subsection{Conditional Erasure}

Models reason about hypothetical Russian success but never about Ukrainian success:

\begin{table}[t]
\centering
\small
\begin{tabular}{lccccr}
\toprule
\textbf{Pattern} & \textbf{GPT} & \textbf{Fl.} & \textbf{Pro} & \textbf{3.1*} & \textbf{Tot.} \\
\midrule
``If R succeeds\ldots'' & 6 & 13 & 4 & 37 & 60 \\
``If R fails\ldots'' & 0 & 0 & 2 & 0 & 2 \\
``If U succeeds\ldots'' & 0 & 0 & 0 & 0 & \textbf{0} \\
``If U fails\ldots'' & 1 & 3 & 0 & 1 & 5 \\
\bottomrule
\end{tabular}
\caption{Conditional framing in reasoning traces. Ukrainian success is never considered as a scenario.}
\label{tab:conditional}
\end{table}

Pro~3.1* produces 37 ``if Russia succeeds'' constructions -- the most of any model -- despite knowing most such outcomes do not occur.

\subsection{Epistemic Framing}

Russia ``claims'' (224 instances) while Ukraine ``denies'' (20 instances; Russia: 1). Ukraine's primary epistemic role is refuting Russian assertions rather than making its own. Russia is also ``confirmed'' 2$\times$ more than Ukraine, creating a paradox where Russian assertions are simultaneously more doubted and more validated.

\subsection{Syntactic Subordination}

``Despite Ukrainian resistance, Russia continues to advance'' appears 30 times. The inverse --``Despite Russian challenges, Ukraine holds'' -- appears 6 times. Ukrainian action is systematically placed in concessive clauses; Russian action occupies the main clause. Ukrainian defense is framed as \emph{overcome}; Russian offense as \emph{persisting}.

\subsection{Composite Scorecard}

\begin{table}[t]
\centering
\small
\begin{tabular}{lcccc}
\toprule
\textbf{Metric} & \textbf{GPT} & \textbf{Flash} & \textbf{Pro} & \textbf{3.1*} \\
\midrule
Mention ratio (R/U) & 1.56 & 1.67 & 1.76 & 1.68 \\
Offensive verb ratio & 2.28 & 3.20 & 3.70 & 3.34 \\
Russia S/F ratio & 8.1 & 9.0 & 6.8 & \textbf{14.1} \\
Ukraine S/F ratio & 1.4 & 2.2 & 2.5 & 1.6 \\
Subject ratio (R/U) & 1.39 & 1.96 & 1.87 & 1.47 \\
Russia first (\%) & 78.6 & 79.6 & 84.0 & \textbf{85.9} \\
``If R succeeds'' & 6 & 13 & 4 & \textbf{37} \\
``If U succeeds'' & 0 & 0 & 0 & 0 \\
\bottomrule
\end{tabular}
\caption{Composite framing scorecard across all four models. Every metric shows the same direction. The contaminated model (3.1*) amplifies framing despite knowing outcomes.}
\label{tab:scorecard}
\end{table}

Every dimension -- agency, verb semantics, success framing, conditional reasoning, epistemic credibility, syntactic structure -- points in the same direction across all four models. The contaminated model, which \emph{knows} most territorial outcomes resolve against offense, produces the most extreme framing on several dimensions. This decoupling of framing from knowledge confirms that the offense-dominant pattern is structural -- embedded in how English-trained LLMs construct conflict narratives -- rather than a reflection of model beliefs about outcomes.

\section{Prediction Error by Time to Resolution}
\label{app:time_remaining}

\begin{figure}[t]
\centering
\includegraphics[width=\columnwidth]{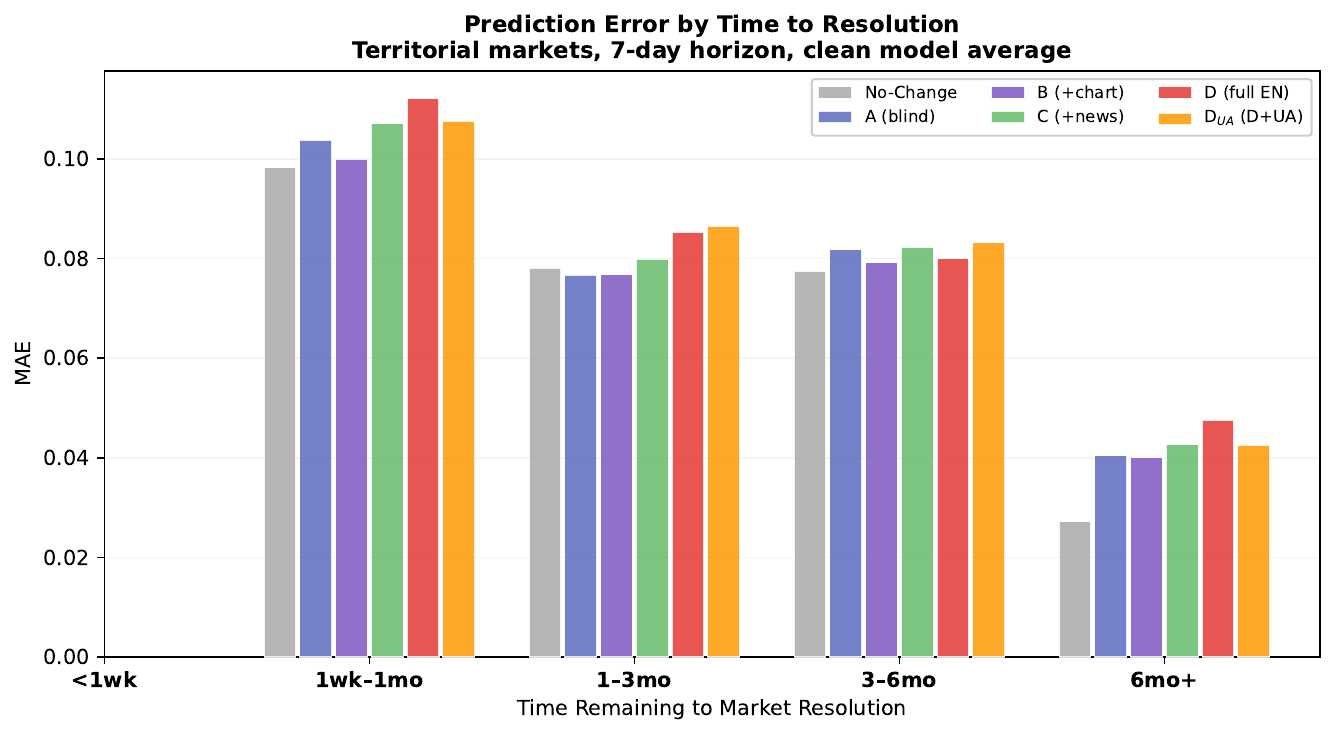}
\caption{MAE by time remaining to market resolution. Context damage is strongest in the 1wk--1mo window where news flow is densest. Territorial markets, 7-day horizon, clean model average.}
\label{fig:time_remaining}
\end{figure}

Figure~\ref{fig:time_remaining} shows prediction error stratified by time remaining until market resolution. Context damage (D exceeding no-change) is concentrated in the 1-week-to-1-month window, where markets are most active and English news coverage is densest. At longer horizons (6+ months), all conditions converge as markets are less liquid and predictions are dominated by the prior.

\section{Case Study: The Zelenskyy Suit}
\label{app:suit}

Consider the Polymarket question ``Will Zelenskyy wear a suit before June?'' All three clean models predicted YES with high confidence (0.91--0.95), reasoning that a papal funeral demands formal attire -- a logically sound inference from generic diplomatic norms. Any Ukrainian would have predicted differently. Zelenskyy has not worn a suit since February 24, 2022; the wartime military clothing is a deliberate political statement, not a wardrobe constraint. Returning to a suit would signal a fundamental shift in how Ukraine frames its wartime posture. The contaminated model, which knows the outcome, predicted 0.33. The gap between 0.93 and 0.33 is not a reasoning failure -- the logic is valid. It is an information ecosystem failure: the English-language corpus encodes ``heads of state wear suits to funerals'' but not ``this particular head of state has made not wearing a suit a defining act of wartime leadership.'' The models reason fluently from the wrong world model.

\end{document}